\documentclass[10pt,twocolumn,letterpaper]{article}

\usepackage[algorithms]{wacv}    
\usepackage[T1]{fontenc}
\usepackage[utf8]{inputenc}
\usepackage{makecell}
\usepackage{multirow}

\definecolor{wacvblue}{rgb}{0.21,0.49,0.74}
\usepackage[pagebackref,breaklinks,colorlinks,allcolors=wacvblue]{hyperref}
\usepackage[accsupp]{axessibility}
\def\wacvPaperID{2836} %
\def\confName{WACV}
\def\confYear{2026}

\title{TaxonRL: Reinforcement Learning with Intermediate Rewards for Interpretable Fine-Grained Visual Reasoning}

\author{Maximilian von Klinski\thanks{Equal contribution} \qquad Maximilian Schall\footnotemark[1]\\
Hasso Plattner Institute\\
University of Potsdam, Germany\\
{\tt\small Maximilian.vonKlinski@student.hpi.de, Maximilian.Schall@hpi.de}
}

\begin{document}
\maketitle
\begin{abstract}
Traditional vision-language models struggle with contrastive fine-grained taxonomic reasoning, particularly when distinguishing between visually similar species within the same genus or family. We introduce \textit{TaxonRL}, a reinforcement learning approach using Group Relative Policy Optimization with intermediate rewards that decomposes the reasoning process into hierarchical taxonomic predictions. Our method incentivizes models to explicitly reason about species-level, genus-level, and family-level features before making final classifications. This structured approach is designed not only to boost accuracy but also to yield a transparent, verifiable decision-making process. On the challenging Birds-to-Words dataset, TaxonRL achieves 91.7\% average accuracy, exceeding human performance (77.3\%) while generating interpretable reasoning traces. We demonstrate strong cross-domain generalization, showing substantial gains in primate and marine species verification. Our results establish that enforcing structured, hierarchical reasoning provides a powerful and transferable framework for fine-grained visual discrimination.
\end{abstract}

\section{Introduction}
\label{sec:intro}

Fine-grained visual recognition has made substantial progress with deep learning, yet distinguishing between closely related subjects remains a challenge. This problem is particularly acute in scientific applications, such as biology, where visually similar species from the same genus require expert-level discrimination \cite{wei_mask-cnn_2016, horn_inaturalist_2018}. Traditional methods like metric learning often produce opaque similarity scores, failing to provide an explanation required for scientific validation and trust \cite{yang_learning_2018, zheng_looking_2019}. Without understanding \textit{why} a decision was made, the applicability of these models in critical domains is severely limited.

Vision-Language Models (VLMs) \cite{radford_learning_2021, liu_visual_2023} present an opportunity to bridge this gap by generating human-readable reasoning. However, standard training paradigms~\cite{ouyang2022traininglanguagemodelsfollow, schulman2017proximalpolicyoptimizationalgorithms, shao_deepseekmath_2024} do not inherently encourage the systematic and hierarchical thinking that experts employ. A model might correctly distinguish two species but do so for the wrong reasons, undermining its reliability. The core challenge, therefore, is not only to improve accuracy, but to instill a logically sound and transparent decision-making process in these models.

\begin{figure*}
    \centering
    \includegraphics[width=\linewidth]{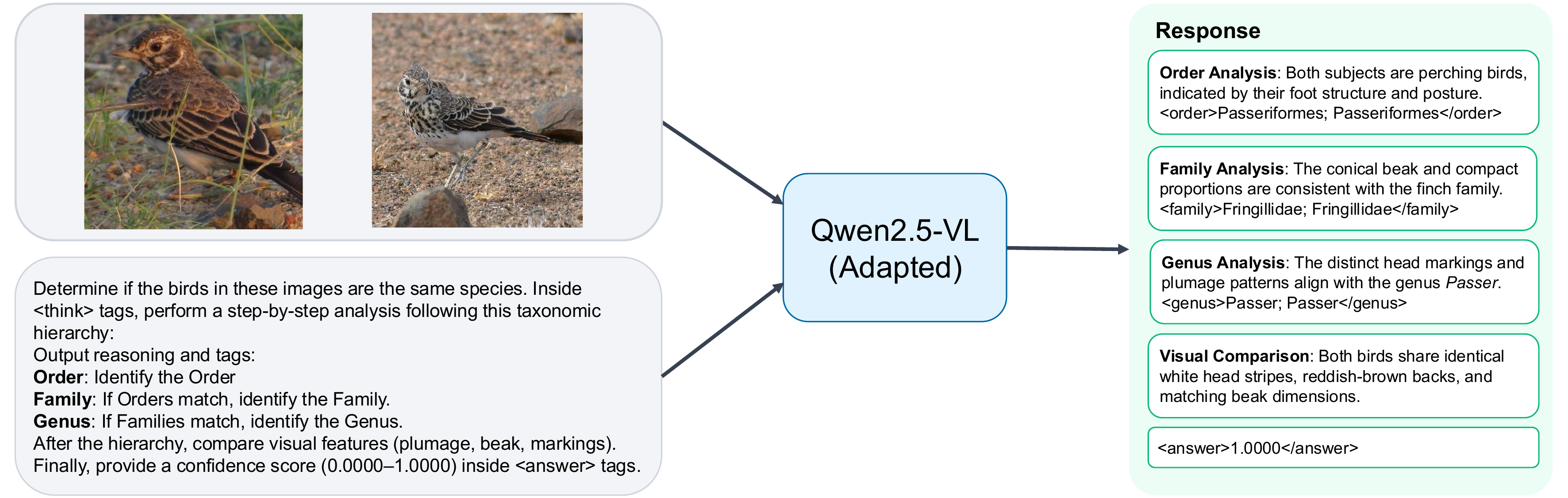}
\caption{\textbf{Hierarchical Reasoning Pipeline.} An example from the Birds-to-Words dataset. The model performs a systematic taxonomic verification (Order $\rightarrow$ Family $\rightarrow$ Genus), grounding each step in visual features to produce an interpretable chain-of-thought and final confidence score.}    
    \label{fig:example_small}
\end{figure*}

We address this by proposing \textit{TaxonRL}, a novel reinforcement learning method that teaches VLMs to reason hierarchically. Our key innovation is an \textbf{intermediate reward mechanism} that decomposes the fine-grained classification task into a sequence of taxonomic predictions (e.g., family, genus, species), as illustrated in \autoref{fig:example_small}. Using Group Relative Policy Optimization (GRPO)~\cite{shao_deepseekmath_2024}, we incentivize the model to generate explicit reasoning traces that identify salient morphological features at each level of the hierarchy before arriving at a final, verifiable conclusion.

We demonstrate that our approach achieves 91.7\% accuracy on the challenging Birds-to-Words~\cite{forbes_neural_2019} dataset, exceeding human performance while generating fully interpretable reasoning. We provide strong evidence of the method's versatility through cross-domain generalization, showing substantial accuracy gains on primate and marine invertebrate verification. Our analysis of the learned reasoning patterns provides insight into the model's decision-making process, establishing that teaching models to reason systematically provides a robust and widely applicable foundation for fine-grained visual discrimination.

Our primary contributions are fourfold. First, we introduce a novel reinforcement learning method that uses an intermediate reward mechanism to enforce hierarchical, step-by-step reasoning in VLMs. Second, our method exceeds human-level performance on the challenging Birds-to-Words dataset, demonstrating its effectiveness in fine-grained species verification. Third, we show the approach's versatility through successful cross-domain generalization to primate and marine species verification tasks. Finally, our approach addresses the "black box" problem by generating interpretable reasoning traces that offer clear, qualitative insight into the model's decision-making process.

\section{Related Work}
\label{sec:relatedwork}

\subsection{Vision-Language Models for Fine-Grained Reasoning}
The advent of VLMs has transformed the landscape of visual understanding and foundational models like CLIP \cite{radford_learning_2021} demonstrated unprecedented zero-shot classification capabilities by aligning images and text in a shared embedding space using contrastive learning. Subsequent architectures, such as BLIP-2 \cite{li_blip-2_2023} and Flamingo \cite{alayrac_flamingo_2022}, improved efficiency and few-shot learning by intelligently connecting powerful pretrained vision encoders with frozen Large Language Models (LLMs). More recent models have focused on enhancing instruction-following and complex reasoning. LLaVA \cite{liu_visual_2023} pioneered the use of visual instruction tuning, while large-scale models—both proprietary like GPT-4o~\cite{openai_gpt-4o_2024} and the Gemini family \cite{team_gemini_2025}, and open-source like Qwen2.5-VL \cite{bai_qwen25-vl_2025}, have demonstrated remarkable capabilities in sophisticated dialogue about images. While these models excel at general visual question answering, they often struggle on fine-grained domains where subtle visual distinctions are critical. 
Our work directly addresses this gap by introducing an explicit training signal to enforce a structured reasoning process.

\subsection{Reinforcement Learning for Model Alignment}

Aligning the outputs of large models with desired behaviors; such as helpfulness, safety, or complex reasoning; is a central challenge. This alignment is typically achieved through a two-stage process: an initial phase of Supervised Fine-Tuning (SFT)~\cite{ouyang2022traininglanguagemodelsfollow} to teach the model the desired output format, followed by RL to refine the reasoning quality \cite{ouyang_training_2022, rafailov_direct_2024, shao_deepseekmath_2024}.

For the RL stage, Reinforcement Learning from Human Feedback (RLHF)~\cite{ouyang_training_2022} was the standard paradigm. To simplify the complex RLHF pipeline, methods like Direct Preference Optimization (DPO)~\cite{rafailov_direct_2024}, optimize the policy directly on preference data, thereby eliminating the need for an explicit reward model.

However, defining a precise reward for multi-step reasoning remains difficult. Group Relative Policy Optimization, introduced by \citet{shao_deepseekmath_2024}, samples multiple responses and calculates relative rewards based on their correctness, thus avoiding an external value function. This paradigm has recently been extended to the multimodal domain. Recent studies like Visual-RFT \cite{liu_visual-rft_2025} and Reason-RFT \cite{tan_reason-rft_2025}, among others \cite{liang_group_2025}, have shown GRPO's effectiveness in fine-tuning VLMs. Our work builds on this line of work by designing a novel intermediate reward structure within the GRPO framework, specifically tailored to guide the model through a hierarchical decision-making process.

\subsection{Classification and Verification }
Fine-Grained discriminative visual classification is a long-standing computer vision challenge focused on distinguishing between subordinate categories, such as bird species \cite{forbes_neural_2019} or car models. Beyond category-level recognition, fine-grained \textit{verification} determines if two images depict the same specific individual. This capability is particularly critical in wildlife conservation for individual re-identification~\cite{freytag_chimpanzee_2016, brust_towards_2017, Brookes2022EvaluatingCE, adam_seaturtleid2022_2024, laskowski2023gorillavision}.
Early deep learning approaches focused on localizing discriminative parts through part annotation \cite{branson_bird_2014}, attention mechanisms \cite{zheng_looking_2019}, region-based learning \cite{chen_destruction_2019}, or part-based models \cite{wei_mask-cnn_2016}, while metric learning aimed to learn embedding spaces where same-class images are closer than different-class images \cite{yang_learning_2018}. While effective in improving accuracy, these methods produce opaque similarity scores.

While some work has explored generating post-hoc explanations for model decisions \cite{selvaraju_grad-cam_2020, achtibat2024attnlrpattentionawarelayerwiserelevance, shrikumar2019learningimportantfeaturespropagating}, we integrate hierarchical reasoning directly into the training loop via intermediate rewards.

\section{Hierarchical Reward Design}

We formulate discriminative fine-grained taxonomic classification as a pairwise verification task. Given two images $I_1$ and $I_2$, the model must determine whether they depict the same species while generating explicit reasoning traces.

Our key contribution is decomposing the reward signal across multiple granularities. We introduce three complementary reward components that collectively guide the model toward structured, accurate, and interpretable predictions.

\noindent \textbf{1. Structure Reward:} A binary reward ensuring adherence to the required output structure:
\begin{equation}
r_{\text{struct}} = \begin{cases}
1, & \text{if output follows format} \\
0, & \text{otherwise}
\end{cases}
\end{equation}

\noindent \textbf{2. Correctness Reward:} We use the negative cross-entropy for final species-level prediction:
\begin{equation}
r_{\text{corr}} = -[y \log(\hat{y}) + (1-y) \log(1-\hat{y})]
\end{equation}
where $y$ is the ground truth label and $\hat{y}$ is the predicted probability. This maintains competitive performance on the primary classification task.

 \noindent\textbf{3. Intermediate Attribute Reward:} A dense reward for correctly predicting $K$ intermediate attributes (e.g., morphological features):
\begin{equation}
r_{\text{attr}} = \frac{1}{K} \sum_{k=1}^{K} \mathbf{1}\{\hat{z}_k = z_k\}
\label{eq:intermediate}
\end{equation}
where $\hat{z}_k$ and $z_k$ are the predicted and ground truth values for the $k$-th attribute. This promotes faithful reasoning by encouraging the model to ground predictions in observable taxonomic features.

\textbf{Total Reward.} The combined reward balances final accuracy with reasoning quality:
\begin{equation}
\begin{aligned}
r_{\text{total}} &= \lambda \cdot r_{\text{struct}} 
&+ \frac{1-\lambda}{2} \cdot r_{\text{corr}} + \frac{1-\lambda}{2} \cdot r_{\text{attr}}
\end{aligned}
\label{fig:total}
\end{equation}
We set $\lambda = 0.4$ to strictly enforce format consistency while weighting reasoning and accuracy equally.

\section{Contrastive Hierarchical Classification}
\label{sec:contrastive_hierarchical_classification}
To validate our hierarchical reasoning approach, we evaluate TaxonRL on contrastive fine-grained verification. This task requires discriminating between highly similar visual categories, making it an ideal testbed for assessing the efficacy of our intermediate reward mechanism.

\subsection{Datasets}
We primarily evaluate on the \textbf{Birds-to-Words}~\cite{forbes_neural_2019} dataset, where taxonomic classifications are derived from the Global Biodiversity Information Facility (GBIF) database~\cite{iNaturalist2025}. The dataset comprises 3,166 image pairs spanning 775 species orders, stratified across varying levels of taxonomic similarity: \textit{Same Species}, \textit{Same Genus}, \textit{Same Family}, \textit{Same Order}, \textit{Same Class}, and \textit{Visual} (visually similar but taxonomically distant). This stratification allows us to pinpoint precisely where the model excels or fails within the biological hierarchy.

To assess generalization beyond birds, we additionally evaluate on the \textbf{Danish Fungi 2020}~\cite{Picek_2022} dataset. This comprehensive dataset contains 51,400 image pairs covering 182 different species, providing a robust testbed for analyzing the transferability of our hierarchical reasoning approach to disjoint biological domains.

We split all datasets with a 70\%/15\%/15\% partition for training, validation, and testing respectively.

\subsection{Supervised Fine-Tuning Dataset}
\label{sec:sft_dataset}
For evaluation with supervised fine-tuning, we generate a synthetic dataset of expert-level reasoning traces using \textit{Gemini 2.0} \cite{team_gemini_2025}. For each image pair in the training set, the teacher model is prompted to generate step-by-step reasoning utilizing the ground-truth hierarchy and visual evidence (e.g., plumage patterns, beak morphology). Due to the significant computational cost of generating synthetic traces and the limited performance improvements observed from SFT alone, we restricted this data generation process exclusively to the Birds-to-Words dataset.

\subsection{Implementation Details}
We use \textit{Qwen2.5-VL-7B-Instruct}~\cite{bai_qwen25-vl_2025} as our VLM backbone. We apply GRPO directly to the pretrained model. We generate $n=16$ rollouts for each prompt, sampled with a temperature of 1.0 and top-$p$ of 1.0 to ensure diversity. We set the format adherence weight $\lambda=0.4$ and use a KL penalty coefficient of $1.0 \times 10^{-2}$ to stabilize training.

We follow the Qwen2.5 output schema, where the model generates reasoning within \texttt{<think>} tags, incorporating specific delimiters for \texttt{<order>}, \texttt{<family>}, and \texttt{<genus>}, followed by the final decision in \texttt{<answer>}.

The model is trained using the AdamW optimizer with a learning rate of $1.0 \times 10^{-6}$ and weight decay of $1.0 \times 10^{-2}$. We employ a cosine learning rate scheduler without warmup and train for 60 epochs with a global batch size of 96. All parameters are updated during training. Experiments were conducted on 6 NVIDIA H100 GPUs.

During evaluation, we use greedy decoding (temperature=0.0) to ensure deterministic predictions. Taxonomic predictions are parsed from the output. The final verification score is thresholded at 0.5 for accuracy computation.

Contrary to recent two-stage approaches~\cite{shao_deepseekmath_2024}, preliminary experiments indicated that SFT as a preliminary stage before RL did not yield significant improvements for this task. We therefore apply GRPO tuning directly to the base pretrained model.

\subsection{Baselines} 
We compare TaxonRL against several established methods. \textit{Neural Naturalist} \cite{forbes_neural_2019} serves as the original benchmark utilizing localized descriptions. We further evaluate \textit{DinoV2\textsubscript{Giant}} \cite{oquab2024dinov2learningrobustvisual} as a strong visual baseline, employing the frozen backbone with separate linear classifiers trained for each taxonomic level.
We also evaluate the basemodel \textit{Qwen2.5-VL-7B} in a zero-shot setting. To isolate the impact of our specific contributions, we include an \textit{SFT-Only} baseline fine-tuned on the synthetic expert traces described in \autoref{sec:sft_dataset}, and a \textit{Standard GRPO} baseline optimized solely on the final verification accuracy.

\section{Results}

\autoref{tab:combined_results} presents our main findings, comparing TaxonRL against both standard baselines and human performance. Beyond the top-line metrics, we provide a deeper look into the model's reasoning process and failure modes to understand where the performance gains actually come from.

\subsection{Main Results}

\textbf{Overall Performance.} Our method achieves 91.7\% average accuracy, exceeding human performance (77.3\%) by 14.4 percentage points and establishing a new state-of-the-art on this benchmark.

To contextualize our approach, we include an SFT-only baseline trained on synthetic expert demonstrations generated by Gemini 2.0. This baseline achieves 72.8\% average accuracy, representing only a modest 1.9 percentage point improvement over the base Qwen2.5-VL-7B model (70.9\%). This marginal gain demonstrates that supervised imitation of reasoning traces alone is insufficient. The model can learn the \emph{format} of hierarchical reasoning but fails to internalize the underlying discriminative strategy.

\textbf{Impact of Reinforcement Learning.} The introduction of GRPO yields substantial improvements. Standard GRPO, optimized solely on final verification accuracy without intermediate rewards, achieves 89.8\% average accuracy, a 17.0 point gain over the SFT baseline. This demonstrates that reward-based optimization is far more effective than supervised learning for this task.

Our full method, incorporating intermediate rewards for hierarchical taxonomic predictions, achieves 91.7\% accuracy, a 1.9 point improvement over standard GRPO. While this gain may appear modest in aggregate, it is highly significant for the most challenging categories. For visually similar but taxonomically distant pairs (the ``Visual'' category), our method achieves 79.4\% versus 72.1\% for standard GRPO, representing a 26.2\% reduction in error rate. This suggests our intermediate reward mechanism successfully enforces more robust reasoning patterns that generalize beyond superficial visual similarity.

\textbf{Performance Across Taxonomic Hierarchy.} Our method exhibits perfect accuracy (100.0\%) for pairs differing at the order, family or genus level, matching GRPO without intermediate rewards and closely outperforming human performance (100\%, 98.0\%, 98.4\%). The key differentiator emerges at finer granularities: for same-genus pairs (different species), we achieve 91.7\% accuracy, and for same-species verification, 83.7\%. These results suggest that our intermediate reward mechanism successfully enforces a more robust and faithful reasoning process. By explicitly incentivizing the model to follow the taxonomic hierarchy, our method grounds its final decision in a verifiable, step-by-step analysis, leading to state-of-the-art, human-level performance. 

\textbf{Generalization to Fungi.} TaxonRL reached 86.9\% accuracy, beating both the base model (70.2\%) and the standard GRPO baseline (82.9\%). This confirms that the structured reasoning learned by TaxonRL is not an artifact of the bird domain, but a transferable skill applicable to disjoint biological hierarchies.

\begin{table*}[ht!]
\centering
\small
\setlength{\tabcolsep}{4pt}
\begin{tabular}{llccccccc}
\toprule
\textbf{Dataset} & \textbf{Model} & \textbf{Visual} & \textbf{\makecell{Same \\ Species}} & \textbf{\makecell{Same \\ Genus}} & \textbf{\makecell{Same \\ Family}} & \textbf{\makecell{Same \\ Order}} & \textbf{\makecell{Same \\ Class}} & \textbf{Average} \\
\midrule
\multirow{7}{*}{\textbf{Bird}} 
 & Neural Naturalist & 55.0 & 45.0 & 67.5 & 70.0 & 72.5 & 77.5 & 64.6 \\
 & DinoV2\textsubscript{Giant} & 52.7 & 69.3 & 59.1 & 70.3 & 85.9 & 100.0 & 76.9 \\
 & Human & 52.5 & 40.2 & 88.7 & 98.4 & 98.0 & \textbf{100.0} & 77.3 \\
 & Qwen2.5-VL-7B & 51.4 & \textbf{86.8} & 41.5 & 70.9 & 90.7 & 93.8 & 70.9 \\
 & SFT & 52.9 & 85.7 & 43.8 & 75.5 & 92.2 & 95.5 & 72.8 \\
 & GRPO & 72.1 & 83.7 & 89.6 & \textbf{100.0} & \textbf{100.0} & \textbf{100.0} & 89.8 \\
 & \textbf{TaxonRL} & \textbf{79.4} & 83.7 & \textbf{91.7} & \textbf{100.0} & \textbf{100.0} & \textbf{100.0} & \textbf{91.7} \\
\midrule
\multirow{4}{*}{\textbf{Fungi}} 
& DinoV2\textsubscript{Giant} & -- & 59.9 & 60.4 & -- & 72.8 & 78.5 & 67.9 \\
 & Qwen2.5-VL-7B & -- & 70.2 & \textbf{61.1} & -- & 84.0 & 85.9 & 74.4 \\
 & GRPO & -- & 77.4 & 60.1 & -- & \textbf{96.5} & \textbf{96.1} & 81.8 \\
 & \textbf{TaxonRL} & -- & \textbf{91.2} & 59.3 & -- & 94.7 & 95.5 & \textbf{88.2} \\
\bottomrule
\end{tabular}
\caption{Performance comparison on Birds-to-Words and Fungi datasets. For the Bird dataset, we report pairwise verification accuracy (\%) across different taxonomic levels. For the Fungi dataset, we report the overall accuracy (\%). Bold values indicate the best performance in each column/category.}
\label{tab:combined_results}
\vspace{-3.6mm}
\end{table*}

\subsection{Analysis of Reasoning Traces}
\label{sec:analysis}

A key advantage of our method is the generation of interpretable reasoning traces, which reveal a clear qualitative improvement over a baseline trained with standard GRPO. The baseline model typically offers holistic visual summaries, while our method produces a structured, hierarchical breakdown of its analysis.

This is evident when comparing a bee-eater and a kingfisher. The baseline's reasoning is a brief summary of visual differences. In contrast, our model provides a more rigorous, expert-like deduction by following the taxonomic hierarchy, as shown in the shortened traces below.

\noindent\textit{Baseline (Standard GRPO):}
\begin{quote}
\small
\texttt{The two birds... are visually distinct... one is characteristic of a bee-eater... the other... a kingfisher.}
\end{quote}
\noindent\textit{Our Method (w/ Intermediate Rewards):}
\begin{quote}
\small
\texttt{<think>}\\
\texttt{1. Order Analysis: ...in the same order.}\\
\texttt{   <order>Coraciiformes; Coraciiformes</order>}\\
\texttt{2. Family Analysis: ...in different families.}\\
\texttt{   <family>Meropidae; Alcedinidae</family>}\\
\texttt{...}\\
\texttt{</think>}
\end{quote}

By first identifying the shared order before pinpointing the crucial difference at the family level, our model provides a more robust and verifiable explanation for its conclusion. This structured process ensures the model identifies salient, fine-grained features, moving beyond superficial similarity. Full, unabridged reasoning traces for this and other examples are provided in the supplementary materials.

\subsection{Quantitative Analysis of Reasoning Traces}
While the qualitative improvements in reasoning structure are evident from inspection, we provide comprehensive quantitative analysis to validate that our intermediate reward mechanism induces genuinely accurate hierarchical reasoning rather than superficial format adherence.

\noindent\textbf{Intermediate Prediction Accuracy:} 
We evaluate the correctness of intermediate taxonomic predictions at each hierarchical level by comparing the model's extracted tags against ground-truth taxonomic labels. \autoref{tab:intermediate_accuracy} presents these results stratified by taxonomic distance between image pairs (Visual category pairs are redistributed among the actual taxonomic differences). Notably, predictions that correctly terminate early (e.g., stopping at the family level when birds belong to different families) are therefore counted as correct on the lower levels.

Our method achieves remarkably high intermediate prediction accuracy across all hierarchical levels, with average accuracies of 97.9\%, 90.1\%, and 86.9\% for order, family, and genus predictions respectively. Performance degrades gracefully at finer taxonomic granularities, which is expected given the increasing visual similarity at lower levels of the hierarchy.

Crucially, these high intermediate accuracies are achieved despite significant distributional shift: 40.65\% of test samples contain genera completely unseen during training, and 94.08\% contain novel species. This demonstrates that the model has learned robust, transferable taxonomic reasoning rather than memorizing training examples. The strong correlation between intermediate prediction accuracy and final answer accuracy (91.7\% average) further validates that the reasoning traces are causally related to the model's decisions rather than post-hoc rationalizations.

\begin{table}[t]
\centering
\small
\begin{tabular}{lcccc}
\toprule
\textbf{Taxonomic Distance} & \textbf{Order} & \textbf{Family} & \textbf{Genus} & \textbf{Answer} \\
\midrule
Same Class     & 100 & 100 & 100 & 100 \\
Same Order     & 99.2 & 99.2 & 99.2 & 98.4 \\
Same Family    & 98.2 & 85.1 & 99.1 & 98.2 \\
Same Genus     & 94.6 & 79.5 & 71.4 & 76.8 \\
Same Species   & 97.8 & 88.7 & 74.2 & 88.2 \\
\midrule
\textbf{Average} & \textbf{97.9} & \textbf{90.1} & \textbf{86.9} & \textbf{91.7} \\
\bottomrule
\end{tabular}
\caption{Intermediate prediction accuracy (\%) at each taxonomic level, stratified by ground-truth taxonomic distance. The model demonstrates high fidelity in hierarchical reasoning across all granularities, with graceful degradation at finer levels.}
\label{tab:intermediate_accuracy}
\end{table}

\noindent\textbf{Format Adherence:} 
All models fine-tuned with reinforcement learning achieve perfect format adherence (100\%), successfully generating valid XML-structured outputs with properly nested tags in all test cases. The base Qwen2.5-VL model achieves 99.3\% format adherence alone. The small remaining format errors are eliminated through either supervised fine-tuning or RL optimization, confirming that format consistency is readily achievable and does not represent a bottleneck for our approach.

\noindent\textbf{Reasoning Trace Length Analysis:}
\autoref{tab:output_length} presents token-level statistics for model outputs across different training paradigms. Our method with concrete intermediate rewards produces substantially longer responses (319 tokens) compared to standard GRPO (121 tokens), reflecting the explicit articulation of hierarchical reasoning at multiple taxonomic levels. Interestingly, SFT produces responses of comparable length (351 tokens) to our method, yet achieves dramatically lower accuracy (72.8\% vs 91.7\%), demonstrating that verbosity alone does not guarantee reasoning quality. This indicates that our approach effectively leverages test-time scaling, where increased token generation corresponds to genuine computational depth rather than the verbosity observed in SFT.
The token overhead introduced by structured reasoning exchanges computational efficiency for interpretability and accuracy. The performance gains (1.9 points over standard GRPO) justify this cost, particularly in scientific domains where transparent decision-making is required.

\begin{table}[t]
\centering
\small
\begin{tabular}{lccc}
\toprule
\textbf{Method} & \textbf{Min} & \textbf{Mean} & \textbf{Max} \\
\midrule
Qwen2.5-VL                        & 72  & 126.53 & 247 \\
SFT                               & 137 & 351.26 & 543 \\
GRPO w/o Intermediate Rewards    & 73  & 120.59 & 260 \\
TaxonRL  & 121 & 319.24 & 537 \\
\bottomrule
\end{tabular}
\caption{Output length statistics (in tokens) across methods. Structured reasoning with concrete intermediate rewards produces longer traces that encode explicit hierarchical analysis, contrasting with the terser outputs of standard GRPO.}
\label{tab:output_length}
\end{table}

\textbf{Failure Mode Analysis.}
False positives were dominated by insufficient morphological discriminability at fine taxonomic granularities (80.0\%), with the remaining cases, especially at larger taxonomic distances, attributable to strong lighting changes or partial occlusion, while false negatives were primarily driven by extreme imaging variations, such as occlusion, lighting artifacts and perspective distortion (81.8\%), with strong sexual dimorphism accounting for the remaining 18.2\%; example pairs for each failure mode are provided in the supplementary materials.

\section{Generalization to Identity Verification}
\label{sec:verification}
We demonstrate that our hierarchical reasoning framework generalizes to animal identity verification. We adapt the intermediate reward structure to focus on individual characteristics (e.g., age, gender, morphology) rather than taxonomic ranks, encouraging the model to explicitly reason about traits that enable individual discrimination.

\subsection{Datasets and Evaluation}
We evaluate our approach on three diverse datasets covering primate and marine species. We formulate the task as \textbf{open-set pairwise verification}, where the identities in the test set are disjoint from those in the training set.  We employ a 70\%/15\%/15\% split for training, validation, and testing. We maintain the same training and evaluation parameters as described in \autoref{sec:contrastive_hierarchical_classification}.

\noindent \textbf{Gorilla-SPAC-Wild}~\cite{schall2025gorillawatchautomatedinthewildgorilla} We leverage this large-scale dataset of Western Lowland Gorillas for re-identification in the wild, generating \textbf{178,570} image pairs across 108 distinct individuals. We focus on the body subset, as it presents significant challenges including extreme pose variation, partial occlusion by vegetation, and dynamic lighting conditions. To support our hierarchical reasoning approach, we utilize five biologically distinct age-sex classes as intermediate prediction targets: \textit{Silverback}, \textit{Adult Female}, \textit{Blackback}, \textit{Adolescent/Juvenile}, and \textit{Infant}.

\noindent \textbf{ChimpFace~\cite{freytag_chimpanzee_2016}} As a standard benchmark for primate facial recognition, this dataset contains \textbf{5,500} pairs derived from 90 individuals. We extract age-group classifications from the metadata (e.g., \textit{Infant}, \textit{Adult}, \textit{Elderly}) to serve as intermediate attributes. This guides the model to decompose identity verification by first analyzing age-related facial structure and pigmentation before assessing individual identity.

\noindent \textbf{SeaStar~\cite{seastar}} To assess generalization to non-mammalian morphologies, we utilize a dataset of \textbf{51,400} pairs from 95 individuals. Sea stars are characterized by radially symmetric body plans, and identification requires analyzing unique surface pigmentation patterns rather than facial geometries.

\subsection{Results}
As shown in \autoref{tab:verifiation_results}, our method significantly outperforms baselines across all three datasets. On the Gorilla dataset, our approach achieves \textbf{78.2\%} accuracy, far surpassing standard GRPO (71.2\%). The gains are even more pronounced on ChimpFace, where our method reaches \textbf{87.4\%}, a substantial improvement over standard GRPO (78.6\%).

On the SeaStar dataset, our method achieves \textbf{95.6\%} accuracy, representing a 1.7 percentage point improvement over standard GRPO (93.9\%). These results are particularly significant given the unique morphological challenges of seastar re-identification. While the high baseline GRPO performance (93.9\%) suggests that verification on this dataset may be comparatively easier than primate identification, possibly due to more distinctive surface patterns, the improvement from intermediate rewards demonstrates that structured reasoning provides value even in relatively easy datasets.

These results confirm that guiding the model to first identify key biological characteristics provides a robust and transferable foundation for identity verification across dramatically different organisms.

\begin{table}[ht!]
    \centering
        \begin{tabular}{l l c}
            \toprule
            \textbf{Dataset} & 
            \textbf{Method} & 
            \textbf{\makecell{Accuracy (\%) \\ (Open-Set)}} \\
            \midrule

            \multirow{4}{*}{\textbf{Gorilla}} & Qwen2.5-VL-7B & 50.2  \\
             & DinoV2\textsubscript{Giant} & 57.6  \\
             & GRPO         & 71.2  \\
            & \textbf{TaxonRL}  & \textbf{78.2}  \\
            \midrule
            \multirow{4}{*}{\textbf{ChimpFace}} & Qwen2.5-VL-7B & 50.3  \\
             & DinoV2\textsubscript{Giant} & 58.1  \\
             & GRPO       & 78.6  \\
             & \textbf{TaxonRL} & \textbf{87.4} \\
            \midrule
            \multirow{4}{*}{\textbf{SeaStar}} & Qwen2.5-VL-7B & 65.7  \\
             & DinoV2\textsubscript{Giant} & 65.4  \\
             & GRPO       & 93.9  \\
             & \textbf{TaxonRL} & \textbf{95.6} \\
            
            \bottomrule
        \end{tabular}
\caption{Open-set identity verification accuracy (\%) across three diverse animal verification datasets.}

    \label{tab:verifiation_results}
\end{table}

\subsubsection*{Example: Focusing on Core Biological Features}
In this case, two images of the same silverback are presented. Our model correctly verifies the match by identifying the key biological marker, while the baseline, failing to recognize this marker, defaults to an incorrect rejection.

\vspace{5pt}
\noindent\textit{Baseline (Standard GRPO) - Incorrect:}
\begin{quote}
\small
\texttt{<think> Both images depict a large brown gorilla... similar in figure and visible features... There are no distinctive marks or identifiers visible that suggest a differ... </think>}
\texttt{<answer>0.0000</answer>}
\end{quote}

\noindent\textit{Our Method (w/ Intermediate Rewards) - Correct:}
\begin{quote}
\small
\texttt{<think>}\\
\texttt{<type>Silverback; Silverback</type>}\\
\texttt{Both images show a large gorilla with a prominent silverback mane, which is a characteristic feature of a Silverback gorilla... There are no visible differences... that would suggest these are two different gorillas.}\\
\texttt{</think>}\\
\texttt{<answer>1.0000</answer>}
\end{quote}

Here, the baseline's reasoning is consistent with its output: it fails to find a unique positive identifier and therefore concludes there is no match. This is a common failure mode for simple similarity models. Our model, in contrast, is guided to first identify the subject's biological type. By confirming both images show a \texttt{<type>Silverback</type>} and recognizing the ``prominent silverback mane,'' it establishes a strong, feature-based justification for its correct positive verification.
We provide more examples in the supplementary materials.

\section{{Concrete vs. Binary Intermediate Labels}}
We investigate whether the model benefits from predicting \emph{concrete} taxonomic labels (e.g., \texttt{<family>Meropidae; Alcedinidae</family>}) or whether \textit{binary} 

predictions at each hierarchical level suffice (e.g., \texttt{<family>different; different</family>}). 

We train an alternative model where the intermediate attribute reward component from \autoref{eq:intermediate} is modified to supervise only binary \texttt{same}/\texttt{different} predictions at each taxonomic level, rather than concrete taxonomic names. Specifically, at each hierarchical level $k$, the model predicts whether the two specimens belong to the same taxon at that level. The binary intermediate reward becomes:

\begin{equation}
r_{\text{attr}}^{\text{binary}} = \frac{1}{K} \sum_{k=1}^{K} \mathbf{1}\{\hat{b}_k = b_k\}
\end{equation}

where $b_k \in \{\text{same}, \text{different}\}$ indicates taxonomic agreement at level $k$, and $\hat{b}_k$ is the model's binary prediction. This replaces $r_{\text{attr}}$ in the total reward (\autoref{fig:total}), while all other components ($r_{\text{struct}}$ and $r_{\text{corr}}$) and training parameters remain identical to our main method.

\autoref{tab:ablation_concrete_vs_binary} shows that while both methods achieve identical performance on coarse levels (Order, Family) and same-species verification, concrete intermediate rewards demonstrate superior performance on the most challenging ``Visual'' category, where concrete labels achieve 79.4\% versus 77.9\%. For same-genus pairs, concrete labels attain 91.7\% compared to 89.6\%, a 2.1 percentage point improvement. We hypothesize that predicting ``\texttt{Meropidae}'' versus ``\texttt{Alcedinidae}'' compels the model to reason about the morphological features that define these families (e.g., curved beaks vs. robust bills), rather than merely detecting a generic difference. This structured knowledge appears most valuable for challenging cases where visual similarity conflicts with taxonomic distance (``Visual'') or where fine-grained within-genus discrimination is required. We conclude that concrete taxonomic labels provide a richer learning signal.

\begin{table}[!htbp]
\centering
\small
\setlength{\tabcolsep}{3.5pt}
\begin{tabular}{lcccccc}
\toprule
\textbf{Reward Type} & \textbf{Visual} & \textbf{Species} & \textbf{Genus} & \textbf{Family} & \textbf{Order} & \textbf{Avg.} \\
\midrule
Binary & 77.9 & 83.7 & 89.6 & 100.0 & 100.0 & 91.1 \\
Concrete & 79.4 & 83.7 & 91.7 & 100.0 & 100.0 & 91.7 \\
\midrule
$\Delta$ & +1.5 & 0.0 & +2.1 & 0.0 & 0.0 & +0.6 \\
\bottomrule
\end{tabular}
\caption{Comparing concrete taxonomic labels versus binary same/different predictions. All values are accuracies (\%).}
\label{tab:ablation_concrete_vs_binary}
\end{table}

\section{Conclusion}

In this work, we addressed the dual challenges of accuracy and interpretability in contrastive fine-grained visual reasoning. We introduced \textit{TaxonRL}, a novel reinforcement learning framework that successfully instills a hierarchical, step-by-step reasoning process in Vision-Language Models. By leveraging a multi-granular intermediate reward mechanism, our method guides the model to emulate an expert's systematic analysis, decomposing complex visual discrimination tasks into a sequence of verifiable predictions. Our experiments demonstrate the efficacy of this approach, exceeding human-level performance on the Birds-to-Words dataset while producing transparent reasoning traces that explain the basis of its conclusions.

We demonstrated that this structured reasoning approach is not confined to biological taxonomy. Its successful generalization to primate and marine invertebrate identity verification underscores a broader principle: enforcing a logical, sequential analysis provides a robust foundation for fine-grained discrimination across diverse domains. This represents a significant step towards building systems whose reasoning is not only correct but also transparent and trustworthy.

While promising, our approach has limitations. It currently relies on a predefined reasoning hierarchy, which may not be available for all tasks. Future research could also explore methods for automatically discovering taxonomic hierarchies from unstructured data, thereby increasing the framework's applicability to novel domains. Additionally, while we achieved state-of-the-art results using Qwen2.5-VL-7B, our evaluation was conducted on a single backbone architecture; future work could verify the scalability and transferability of TaxonRL to other VLM architectures.

The application of this framework to identity verification carries inherent ethical risks, particularly regarding potential misuse in surveillance and privacy encroachment. Furthermore, such systems are susceptible to performance biases across demographic groups. However, a key mitigating factor of our approach is its inherent transparency; unlike black-box systems, the generation of an explicit reasoning trace allows for the auditing of model decisions to detect and correct discriminatory reasoning patterns.

We release our code and datasets to facilitate further research and reproducibility at \url{https://github.com/max-vkl/TaxonRL}.

\section*{Acknowledgments}
The project on which this report is based was funded by the Federal Ministry of Research, Technology and Space under the funding code “KI-Servicezentrum Berlin-Brandenburg” 16IS22092. We also acknowledge the support of the Sabine Plattner African Charities for their contribution to this research. Responsibility for the content of this publication remains with the author.

{
    \small
    \bibliographystyle{ieeenat_fullname}
    \bibliography{references_taxonrl_new}
}
\appendix

\end{document}